\documentclass{article}

\usepackage{times}
\usepackage{graphicx} 
\usepackage{subfigure} 
\usepackage{epstopdf}

\usepackage{natbib}

\usepackage{algorithm}
\usepackage{algorithmic}

\usepackage[accepted, nohyperref]{icml2017} 

\usepackage{amsmath,amssymb,amsfonts}

\newcommand{\func}[2]{#1\left(#2\right)}
\newcommand{\ind}[2]{#1^{(#2)}}
\newcommand{\EE}[2]{\mathbb{E}_{ #1 }\left[ #2 \right]}

\newcommand{\iid}{\overset{iid}{\sim }}
\newcommand{\R}{\mathbb{R}}
\newcommand{\calD}{\mathcal{D}}
\newcommand{\calN}{\mathcal{N}}
\newcommand{\ud}{\mathrm{d}}

\icmltitlerunning{The Statistical Recurrent Unit}

\begin{document} 

\twocolumn[
\icmltitle{The Statistical Recurrent Unit}




\begin{icmlauthorlist}
\icmlauthor{Junier B. Oliva}{cmu}
\icmlauthor{Barnab\'{a}s P\'{o}czos}{cmu}
\icmlauthor{Jeff Schneider}{cmu}
\end{icmlauthorlist}

\icmlaffiliation{cmu}{Machine Learning Department, Carnegie Mellon University}

\icmlcorrespondingauthor{Junier B. Oliva}{joliva@cs.cmu.edu}

\icmlkeywords{recurrent neural networks, summary statistics, time-series, sequences}

\vskip 0.3in
]



\printAffiliationsAndNotice{\icmlEqualContribution} 

\begin{abstract} 
Sophisticated gated recurrent neural network architectures like LSTMs and GRUs have been shown to be highly effective in a myriad of applications. We develop an un-gated unit, the statistical recurrent unit (SRU), that is able to learn long term dependencies in data by only keeping moving averages of statistics. The SRU's architecture is simple, un-gated, and contains a comparable number of parameters to LSTMs; yet, SRUs perform favorably to more sophisticated LSTM and GRU alternatives, often outperforming one or both in various tasks. We show the efficacy of SRUs as compared to LSTMs and GRUs in an unbiased manner by optimizing respective architectures' hyperparameters in a Bayesian optimization scheme for both synthetic and real-world tasks.
\end{abstract} 

\section{Introduction}
The analysis of sequential data has long been a staple in machine learning. Domain areas like natural language \cite{zaremba2014recurrent, vinyals2015grammar}, speech \cite{graves2013speech, graves2014towards}, music \cite{chung2014empirical}, and video \cite{donahue2015long} processing have recently garnered much attention. While the study of sequences itself is broad and may be extended to general functional analysis \cite{ramsay2002applied}, most recent success has been from neural network based models, especially from recurrent architectures.

Recurrent networks are dynamical systems that represent time recursively. For example, the simple recurrent unit \cite{elman1990finding} contains a hidden state that itself depends on the previous hidden state. However, training such networks has been observed to be difficult in practice due to exploding and vanishing gradients when propagating error gradients through time \cite{hochreiter2001gradient}. While exploding gradients can be mitigated with techniques like gradient clipping and normalization \cite{pascanu2013difficulty}, vanishing gradients may be harder to deal with. As a result, sophisticated gated architectures like Long-Short Term Memory (LSTM) networks \cite{hochreiter1997long} and Gated Recurrent Units (GRU) networks \cite{cho2014properties} have been developed. These gated architectures contain ``memory cells'' along with gates to control how much they decay through time thereby aiding the networks' ability to learn long term dependencies in sequences.

Notwithstanding, there are still challenges in capturing long term dependencies in gated architectures \cite{le2015simple}. In this paper we present a simple un-gated architecture, the Statistical Recurrent Unit, that often outperforms these more complicated alternatives. Although the SRU keeps only simple moving averages of summary statistics, its novel architecture makes it more adept than previous gated units for capturing long term information in sequences and comparing them across different windows of time. 
For instance, the SRU, unlike traditional recurrent units, can obtain a multitude of viewpoints of the past by simple linear combinations of only a few averages. 
We shall illustrate the efficacy of the SRU below using both real-world and synthetic sequential data tasks.

The structure of the paper is as follows: first we detail the architecture of the SRU as well as provide several key intuitions and insights for its design; after, we describe our experiments comparing the SRU to popular gated alternatives, and we perform a ``dissective'' study of the SRU, gaining further understanding of the unit by exploring how various hyper-parameters affect performance; finally, we discuss conclusions from our study. 

\section{Model}
\label{seq:model}
The SRU maintains long term sequential dependencies in a rather intuitive fashion--through summary statistics.
As the name implies, statisticians often employ summary statistics when trying to represent a dataset.
Quite naturally then, we look to an algorithm that itself learns to represent data seen previously in much the same vein as a neural statistician \cite{edwards2016towards}.

Of course, unlike with unordered i.i.d. samples, simply averaging statistics of sequential points will lose valuable temporal information. The SRU maintains sequential information in two ways: first, we generate recurrent statistics that depend on a context of previously seen data; second, we generate moving averages at several scales, allowing the model to distinguish the type of data seen at different points in the past. We expound on these methods for creating temporally-aware statistics below.

We shall see that the statistical design of the SRU yields a powerful yet simple model that is able to analyze sequential data and, on the fly, create summary statistics for learning over sequences. Furthermore, through the use of ReLUs and exponential moving averages, the SRU is able to overcome vanishing gradient issues that are common to many recurrent units.

\subsection{Recurrent Statistics}

We consider an input sequence of real valued points $x_1, x_2, \ldots, x_T \in \R^d$. As seen in the second row of Table \ref{tbl:seq_stats}, we can compute a vector of statistics $\phi(x_i) \in \R^D$ for each point. Here, each vector $\phi(x_i)$ is independent of other points $x_j$ for $j\neq i$. One may then average these vectors as $\mu = \frac{1}{T} \sum_{i=1}^T \phi(x_i)$ to produce summary statistics of the sequence. This approach amounts to treating the sequence as a set of i.i.d. points drawn form some distribution and marginalizing out time. Clearly, here one will lose temporal information that will be useful for many sequence related ML tasks. It is interesting to note that global average pooling operations have gained a lot of recent traction in convolutional networks \cite{lin2013network, iandola2016squeezenet}. Analogously to the i.i.d. statistic approach, global averaging will lose spatial information, yet the high-level summary statistics provide an effective representation. Still, not marginalizing out time should provide a more robust approach for sequence tasks, thus we consider the following methods for producing statistics.

First, we  provide temporal information whilst still utilizing averages through recurrent statistics that also depend on the values of previous points (see third row of Table \ref{tbl:seq_stats}). That is, we compute our statistics on the $i^{\mathrm{th}}$ point $x_i$ not only as a function of $x_i$, but also as a function of the previous statistics of $x_{i-1}$, $\vec{\gamma}_{i-1}$ (which itself depends on $\vec{\gamma}_{i-2}$, etc.):
\begin{align}
\vec{\gamma}_{1} = \gamma(x_1, \vec{\gamma}_{0}), \
\vec{\gamma}_{2} = \gamma(x_2, \vec{\gamma}_{1}), \
\ldots
\label{eq:gamma}
\end{align}
where $\gamma(\cdot, \cdot)$ is a function for producing statistics given the current point and previous statistics, and $\vec{\gamma}_{0}$ is a constant initial vector for convention. 
We note that from a general standpoint if given a flexible model and enough dimensions, then recurrent summary statistics like \eqref{eq:gamma} can perfectly encode ones sequence. Take for instance the following illustrative example where $x_i \in \R^+$ and statistics
\begin{align}
    \vec{\gamma}_i &= (0, \ldots, 0, T x_i, 0, \ldots) \\
    \vec{\gamma}_{i+1} &= (0, \ldots, 0, 0, T x_{i+1}, 0, \ldots).
\end{align}
That is, one records the $i^{\mathrm{th}}$ input in the $i^{\mathrm{th}}$ index. When averaged the statistics will be $\frac{1}{T}\sum_{i=1}^T\vec{\gamma}_i = (x_1, x_2, \ldots)$, i.e. the complete sequence. Such recurrent statistics will undoubtedly suffer from the curse of dimensionality. Hence, we consider a more restrictive model of recurrent statistics which we expound on below \eqref{eq:stats}.

Second, we provide even more temporal information by considering summary statistics at multiple scales. We shed light on the dynamics of statistics through time by using several weights of the same summary statistics. As a simple hypothetical example consider taking multiple means across separate time windows (for instance taking means over indices 1-10, then over indices 11-20, etc.). Such an approach \eqref{eq:winavgs} will illustrate how summary statistics evolve through time. 
\begin{align}
    \underbrace{\phi_1,\ldots, \phi_{10}}_{\mu_{1:10}},\ 
    \underbrace{\phi_{11},\ldots, \phi_{20}}_{\mu_{11:20}},\ \ldots .\label{eq:winavgs}
\end{align}
The SRU will use exponential moving averages $\mu_i = \alpha \vec{\gamma}_i + (1-\alpha) \mu_{i-1}$ to compute means; hence, we consider multiple weights by taking the exponential means at various scales $\alpha_1, \ldots, \alpha_m$ as shown in the last row of Table \ref{tbl:seq_stats}. Later we show that this multi-scaled approach is capable of a combinatorial number of viewpoints of past statistics through simple linear combinations.

\begin{table}[th]
\caption{Methods for keeping statistics of sequences.}
\label{tbl:seq_stats}
\vskip 0.15in
\begin{center}
\begin{small}
\begin{tabular}{cc}
\hline
\abovespace\belowspace
 inputs & $x_1,\ x_2,\ \ldots,\ x_{T}$  \\  
\hline
\abovespace \belowspace
\begin{tabular}{@{}c@{}}
\abovespace
i.i.d. \\
\belowspace
statistics
\end{tabular}
&  $\phi(x_1),\ \phi(x_2),\ \ldots,\  \phi(x_{T})$ \\  
\hline
\abovespace \belowspace
\begin{tabular}{@{}c@{}}
\abovespace
recurrent\\
\belowspace
statistics
\end{tabular}
&  $\gamma\left(x_1, \vec{\gamma}_0\right),\ \gamma(x_2, \vec{\gamma}_1),\ \ldots, \gamma(x_T, \vec{\gamma}_{T-1})$ \\  
\hline
\begin{tabular}{@{}c@{}}
\abovespace
recurrent\\
multi-scaled\\
\belowspace
statistics
\end{tabular}
&
$
\begin{smallmatrix}
\abovespace
 \alpha_1^{T-1} \gamma(x_1, \vec{\gamma}_{0}), & \alpha_1^{T-2} \gamma(x_2, \vec{\gamma}_{1}), & \ldots\\
 & & \ddots  \\
 \belowspace
 \alpha_m^{T-1} \gamma(x_1, \vec{\gamma}_{0}), & \alpha_m^{T-2} \gamma(x_2, \vec{\gamma}_{1}), & \ldots \\
\end{smallmatrix}
$
\\  
\hline
\end{tabular}
\end{small}
\end{center}
\vskip -0.1in
\end{table}

\subsection{Update Equations}
We have discussed in broad terms how one may create temporally-aware summary statistics through multi-scaled recurrent statistics. Below, we cover specifically how the SRU creates and uses summary statistics for sequences.

Recall that our input is a sequence of ordered points: $\{x_{1}, x_{2}, \ldots\}$, $x_{t} \in \R^{d}$. Throughout, we apply an element-wise non-linearity $\func{f}{\cdot}$, which we take to be the ReLU \cite{jarrett2009best, nair2010rectified}: $\func{f}{\cdot} = \max(\cdot, 0)$.
The SRU operates via exponential moving averages, $\ind{\mu}{\alpha} \in \R^{s}$ \eqref{eq:move_avg}, kept at various scales $\alpha \in A = \{\alpha_1, \ldots, \alpha_m\}$, where $\alpha_{i} \in [0, 1)$. These moving averages, $\ind{\mu}{\alpha}$, are of recurrent statistics $\varphi$ \eqref{eq:stats} that are dependent not only on the current input but also on features of averages, $r$ \eqref{eq:recur_stats}. The moving averages are then concatenated as $\mu = (\ind{\mu}{\alpha_1}, \ldots, \ind{\mu}{\alpha_m})$ and used to create an output $o$ \eqref{eq:out} that is fed upwards in the network.

\begin{figure}[ht]
\vskip -0.0in
\begin{center}
\centerline{\includegraphics[width=0.75\columnwidth]{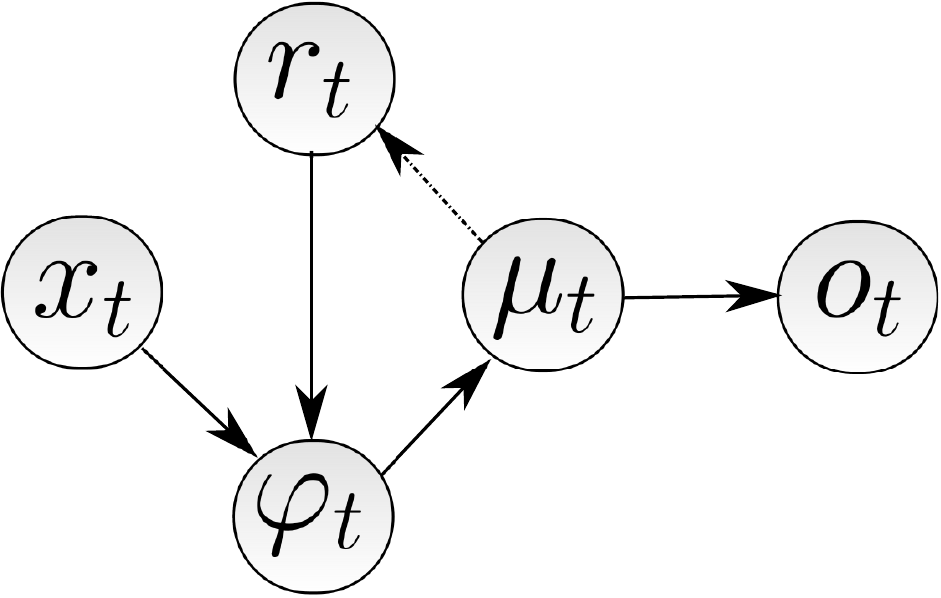}}
\caption{Graphical representation of the SRU. Solid lines indicate a dependence on the current value of a node. Dashed lines indicate a dependence on the previous value of a node. We see that both the current point $x_t$ as well as a summary of the previous data $r_t$ are used to make statistics $\varphi_t$, which in turn are used in moving averages $\mu_t$, finally an output $o_t$ is feed-forward through the rest of the network.}
\label{fig:updates}
\end{center}
\vskip -0.2in
\end{figure} 

We detail the update equations for the SRU below (and in Figure \ref{fig:updates}):
\begin{gather}
r_t = \func{f}{\ind{W}{r} \mu_{t-1} + \ind{b}{r}} \label{eq:recur_stats}\\
\varphi_t = \func{f}{\ind{W}{\varphi} r_t + \ind{W}{x} x_t + \ind{b}{\varphi}} \label{eq:stats}\\
\forall \alpha \in A,\ \ind{\mu}{\alpha}_t = \alpha \ind{\mu}{\alpha}_{t-1} + (1-\alpha) \varphi_t \label{eq:move_avg}\\
o_t = \func{f}{\ind{W}{o} \mu_{t} + \ind{b}{o}} \label{eq:out}.
\end{gather}
In practiced we noted that it suffices to use only a few $\alpha$'s such as $A=\{0, 0.25, 0.5, 0.9, 0.99\}$.

It is worth noting that exponential averages of inputs has been considered previously \cite{mikolov2014learning}. However, that approach performs a moving average of a linear features (specifically the identity mapping) that depends only on the current observation, which is fairly inflexible. Furthermore, such work considers only one scale per feature, limiting the views available per statistic to just one. The use of ReLUs in recurrent units has also been recently explored by \citet{le2015simple}, however there no statistics are kept and their use is limited to the simple RNN when initialized in a special manner.

\subsection{Intuitions from Mean Map Embeddings}
The design of the SRU is deliberately chosen to allow for long term dependencies to be learned. To better elucidate the design and its intuition, let us take a brief excursion to another use of (summary) statistics in machine learning for the representation of data: mean map embeddings (MMEs) of distributions \cite{smola2007hilbert}. At its core, the concept of MMEs is that one may embed, and thereby represent, a distribution through statistics (such as moments). The MME for a distribution $\calD$ given a positive semidefinite kernel $k$ is: 
\begin{equation}
\mu[\calD] = \EE{ X \sim \calD}{\phi_k(X)}, \label{eq:mme_pop}
\end{equation}
where $\phi_{k}$ are the reproducing kernel Hilbert space (RKHS) features of $k$, which may be infinite dimensional. To represent a set $Y = \{y_1, \ldots, y_n\} \iid \calD$ one would use an empirical mean version of the MME:
\begin{equation}
\mu[Y] = \frac{1}{n}\sum_{i=1}^n\phi_k(y_i). \label{eq:mme_samp}
\end{equation}
Numerous works have shown success in representing distributions and sets through MMEs \cite{muandet2016kernel}. One interpretation for the design of SRUs is that we are modifying MME's for use on sequences. Of course, one way of applying MMEs directly on sequences is to simply ignore the non-i.i.d. nature of sequences and treat points as comprising a set. This however loses important sequential information, as previously mentioned. Below we discuss the specific modifications we make from traditional MMEs and the benefits they yield.

\subsubsection{Data-driven Statistics}
First, we note the clear analogue between the mean embedding of a set $Y$, $\mu[Y]$ \eqref{eq:mme_samp}, and the moving average $\ind{\mu}{\alpha}$ \eqref{eq:move_avg}. The moving averages $\ind{\mu}{\alpha}$ are clearly serving as summary statistics of previously seen data. However, the statistics we are averaging for $\ind{\mu}{\alpha}$, $\varphi$ \eqref{eq:stats}, are not comprised of a-priori RKHS features as is typical with MMEs, but rather are learned non-linear features. This has the benefit of using data-driven statistics, and may be interpreted as using a linear kernel in the learned features. 

\subsubsection{Recursive Statistics from the Past}
Second, recall that typical MMEs use statistics that depend only on a single point $x$, $\phi_k(x)$. As aforementioned this is fine for i.i.d. data, however it loses sequential information when averaged. Instead, we wish to assign statistics that depend on the data we have seen so far, since it provides context for one's current point in the sequence. For instance, one may want to have a statistic that keeps track of the difference between the current point and the mean of previous data. We provide a context based on previous data by making the statistics considered at time $t$, $\varphi_t$ \eqref{eq:stats}, a function not only of $x_t$ but also of $\{x_1, \ldots, x_{t-1}\}$ through $r_t$ \eqref{eq:recur_stats}. $r_t$ may be interpreted as a condensation of the sequence seen so far, and allows us to keep sequential information even through an averaging operation. 

\subsubsection{Multi-scaled Statistics}
\begin{figure}[ht]
\vskip -0.0in
\begin{center}
\centerline{\includegraphics[width=1.0\columnwidth]{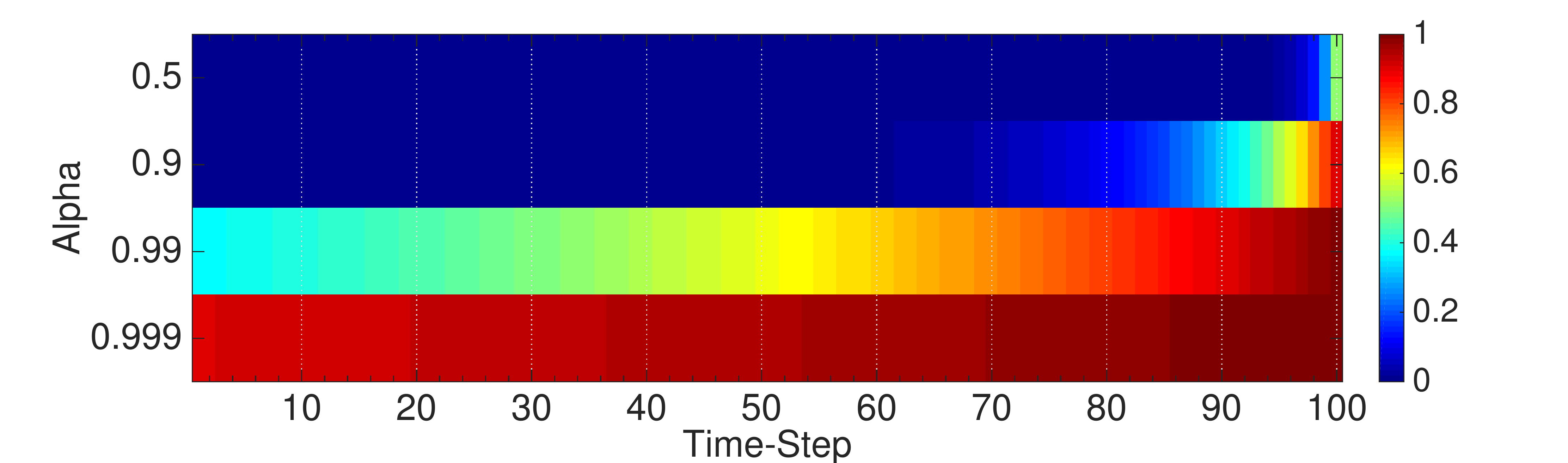}}
\caption{We may unroll the moving average updates as \eqref{eq:unrolled}. To visualize the different emphasis in the past that varying $\alpha$ has on statistics we plot the values of weights in moving averages (i.e. $\alpha^i$) for $100$ points in the past across rows. We see that alpha values closer to $0$ focus only on the recent past, where values close to $1$ maintain an emphasis on the distant past as well.}
\label{fig:alpha_views}
\end{center}
\vskip -0.2in
\end{figure}

Third, the use of multi-scaled moving averages of statistics gives the SRU a simple and powerful rich view of past data that is unique to this recurrent unit. In short, by keeping moving averages at different scales $\{\alpha_1, \ldots, \alpha_m\}$, we are able to uncover differences in statistics at various times in the past. Note that we may unroll moving averages as:
\begin{equation}
\ind{\mu}{\alpha}_t = (1-\alpha)\left( {\varphi}_{t} + \alpha {\varphi}_{t-1} +\alpha^2 {\varphi}_{t-2} + \ldots \right) \label{eq:unrolled}
\end{equation}
Thus, a smaller $\alpha$ weighs current statistics more than older statistics; hence, a concatenated vector $\mu = (\ind{\mu}{\alpha_1}, \ldots, \ind{\mu}{\alpha_m})$ itself provides a multi-scale view of statistics through time (see Figure \ref{fig:alpha_views}). For instance, keeping statistics for short and long terms pasts already yields information on the evolution of the sequence through time.

\subsection{Viewpoints of the Past}
\begin{figure}[ht]
\vskip -0.0in
\begin{center}
\centerline{\includegraphics[width=1.1\columnwidth]{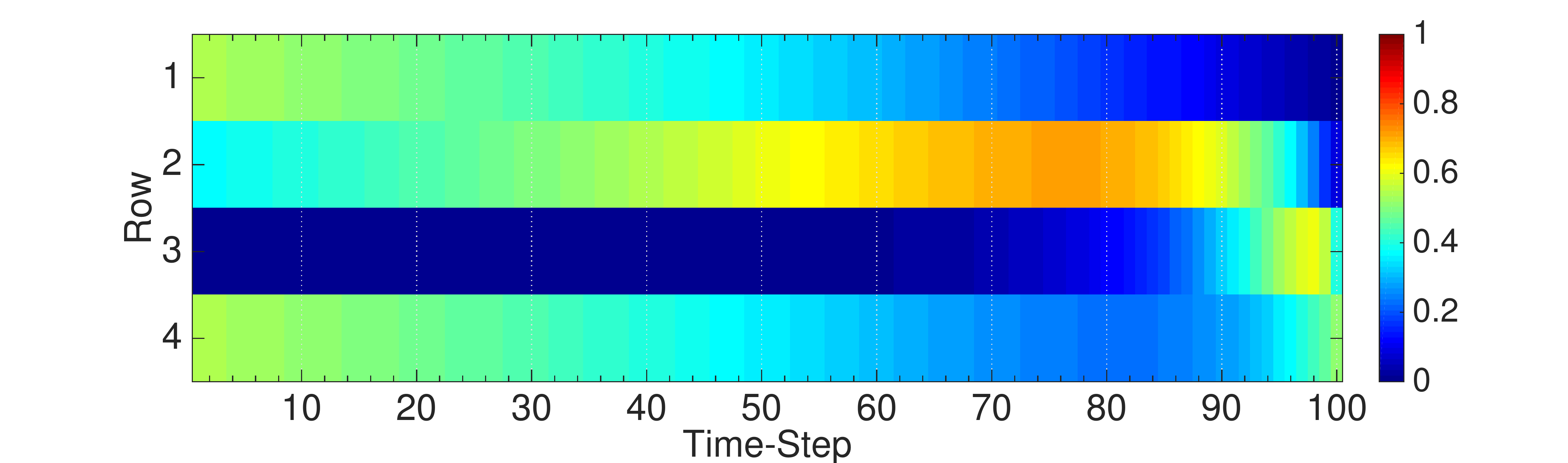}}
\caption{
We visualize the power of taking linear combinations of $\ind{\mu}{\alpha}$'s for providing different viewpoints into past data. In row 1 we show the effective weights that would be used for weighing statistics $\varphi_t$ if one considers $.001^{-1}\ind{\mu}{.999}-.01^{-1}\ind{\mu}{.99}$;
we see that this is equivalent to considering only statistics from the distant past. 
Similarly, we show the effective weights when taking $.01^{-1}\ind{\mu}{.99}-.1^{-1}\ind{\mu}{.9}$ and $.1^{-1}\ind{\mu}{.9}-.5^{-1}\ind{\mu}{.5}$ on rows 2 and 3 respectively. 
We see that these linear combinations amount to considering viewpoints concentrated at various points in the past. Lastly its worth noting that more complicated linear combinations may lead to even richer views on previous statistics; 
for instance, we show $.001^{-1}\ind{\mu}{.999}-.01^{-1}\ind{\mu}{.99}+\frac{.5}{.09}\ind{\mu}{.9}$ 
on row 4, which concentrates on the statistics of the distant and very recent past, but de-emphasizes statistics of data from less recent past.
} 
\label{fig:alpha_lincombos}
\end{center}
\vskip -0.2in
\end{figure}

An interesting and useful property of keeping multiple scales for each statistic is that one can obtain a combinatorial number of viewpoints of the past through simple linear combinations of ones statistics. 
For instance, for properly chosen $w_j, w_k \in \R$, $w_j \ind{\mu}{\alpha_j} - w_k \ind{\mu}{\alpha_k}$ provides an aggregate of statistics from the past for $\alpha_j>\alpha_k$ (Figure \ref{fig:alpha_lincombos}). Of course, more complicated linear combinations may be performed to obtain richer viewpoints that are comprised of multiple windows. Furthermore, by using a linear projection of our statistics $\mu_t$, as we do with $o_t$ \eqref{eq:out}, we are able to compute output features of combined viewpoints of several statistics.

This kind of multi-viewpoint perspective of previously seen data is difficult to produce in traditional gated recurrent units since they must encode where in the sequence they currently are and then store an activation on separate nodes per each viewpoint for future use. SRUs, on the other hand, only need to take simple linear combinations to capture various viewpoints in the past. For example, as shown above, statistics from just the distant past are available via a simple subtraction of two moving averages (Figure \ref{fig:alpha_lincombos}, row 1). Such a windowed view would require a gated unit to learn to stop averaging after a certain point in the sequence, and the corresponding statistic would not yield an information outside of this window. In contrast, each statistic kept by the SRU provides a combinatorial number of varying perspectives in the past through linear combinations and their multi-scaled nature.

\subsection{Vanishing Gradients}
As previously mentioned, it has been shown that vanishing gradients make learning recurrent units difficult due to an inability to propagate error gradients through time. Notwithstanding its simple un-gated structure, the SRU features several safeguards against vanishing gradients. First, units and statistics are comprised of ReLUs. ReLUs have been observed to be easier to train for general deep networks \cite{nair2010rectified} and have had success in recurrent units \cite{le2015simple}. Intuitively, ReLUs allow for the propagation on error on positive inputs without saturation and vanishing gradients as with traditional sigmoid units. The ability of the SRU to use ReLUs (without any special initialization) makes it especially adept at learning long term dependencies through time. 

Furthermore, the explicit moving average of statistics allows for longer term learning. Consider the following derivative of the error signal $E$ w.r.t. an element $\left[ \ind{\mu}{\alpha}_{t-1} \right]_k$ of the unit's moving averages:
\begin{small}
\[
\frac{\ud E}{\ud \left[ \ind{\mu}{\alpha}_{t-1} \right]_k}
= \frac{\ud \left[ \ind{\mu}{\alpha}_{t} \right]_k}{\ud \left[ \ind{\mu}{\alpha}_{t-1} \right]_k}
\frac{\ud E}{\ud \left[ \ind{\mu}{\alpha}_{t} \right]_k}
= \alpha
\frac{\ud E}{\ud \left[ \ind{\mu}{\alpha}_{t} \right]_k}.
\]
\end{small}
That is, the factor $\alpha$ directly controls the decay of the error signal through time and be be easily set to avoid vanishing gradients.

\section{Experiments}
We compared the performance of the SRU to two popular gated recurrent units, the GRU and LSTM unit. All experiments were performed in \texttt{Tensorflow} \cite{abadi2016tensorflow} and used the standard implementations of \texttt{GRUCell} and \texttt{BasicLSTMCell} for GRUs and LSTMs respectively. In order to perform a fair, unbiased comparison of the recurrent units and their hyper-parameters, which greatly affect performance \cite{bergstra2012random}, we used the \texttt{Hyperopt} \cite{bergstra2015hyperopt} Bayesian optimization package. We believe that such an approach gives each algorithm a fair shot to succeed without injecting biases from experimenters or imposing gross restrictions on architectures considered. 

In all experiments we used SGD for optimization using gradient clipping \cite{pascanu2013difficulty} with a norm of $1$ on all algorithms. Unless otherwise specified $100$ trials were performed to search over the following hyper-parameters on a validation set: one, \texttt{initial\_learning\_rate} the initial learning rate used for SGD, in range of $ [\exp(-10), 1]$; two, \texttt{lr\_decay} the multiplier to multiply the learning rate by every $1$k iterations, in range of $[0.8, 0.999]$; three, \texttt{dropout\_keep\_rate}, percent of output units that are kept during dropout, in range $(0, 1]$; four, \texttt{num\_units} number of units for recurrent unit, in $\{1, \ldots, 256\}$. In addition, the following two parameters were searched over for the SRU: \texttt{num\_stats}, the dimensionality of $\varphi$ \eqref{eq:stats}, in $\{1, \ldots, 256\}$; \texttt{summary\_dims}, the dimensionality of $r$ \eqref{eq:recur_stats}, in $\{1, \ldots, 64\}$.

\subsection{Synthetic Recurrent Unit Generated Data}
\label{sec:synth}
First we provide evidence that traditional gated units have difficulties capturing the same type of multi-scale recurrent statistic based dependencies that the SRU offers.
We show the relative inefficiency of traditional gated units at learning long term dependencies of statistics by considering 1d synthetic data from a ground truth SRU. 

We begin the sequences with $x_1 \iid \calN(0, 100^2)$, and $x_t$ is the results of a projection of $o_{t}$. We generate a total of 176 points per sequence for $3200$ training sequences, $400$ validation sequences, and $400$ testing sequences.

The ground truth statistical recurrent unit has three statistics $\phi_t$ \eqref{eq:stats}: the positive part of inputs $(x)_+$, the negative part of inputs $(x)_-$, and an internal statistic, $z$. We use $\alpha \in \{\alpha_i\}_{i=1}^{5} = \{0.0, 0.5, 0.9, 0.99, 0.999\}$. Denote $\ind{\mu}{\alpha}_{+}$, $\ind{\mu}{\alpha}_{-}$, $\ind{\mu}{\alpha}_{z}$ as the moving averages using $\alpha$ for each respective statistic.
The internal statistic $z$ does not get used (through $r_t$ \eqref{eq:recur_stats}) in updating the statistics for $(x)_+$ or $(x)_-$. $z$ is itself updated as:
\begin{small}
\begin{align*}
z_t &= \left(z_{t-1}\right)_{+} \\
&+ \left(\ind{\mu}{\alpha_4}_{+} - \ind{\mu}{\alpha_5}_{+} - 0.01\right)_{+}
 - \left(-\ind{\mu}{\alpha_4}_{-} + \ind{\mu}{\alpha_5}_{-} - 0.01\right)_{+} \\ 
&- \left(-\ind{\mu}{\alpha_4}_{+} + \ind{\mu}{\alpha_5}_{+} - 0.05\right)_{+}
 + \left(\ind{\mu}{\alpha_4}_{-} - \ind{\mu}{\alpha_5}_{-} - 0.05\right)_{+},
\end{align*}
\end{small}
where each of the summands are $r_t$ features.
Furthermore we have $o_t \in \R^{15}$ \eqref{eq:out}:
\[
o_t = \left((x_{t})_+,\ -(x_{t})_-,\ v_1^T \mu_t,\ \ldots,\ v_{13}^T \mu_t\right),
\]
where $v_j$'s where initialized and fixed as $(v_j)_k \iid \calN(0, (\frac{1}{100})^2)$. Finally the next point is generated as:
\[
x_{t+1} = (x_{t})_+ - (x_{t})_- + w^T o_{t,3:} ,
\]
where $w$ was initialized and fixed as $(w)_k \iid \calN(0, 1)$, and $o_{t,3:}$ are the last 13 dimensions of $o_{t}$.

\begin{figure}[ht]
\vskip -0.1in
\begin{center}
\centerline{\includegraphics[width=0.5\columnwidth]{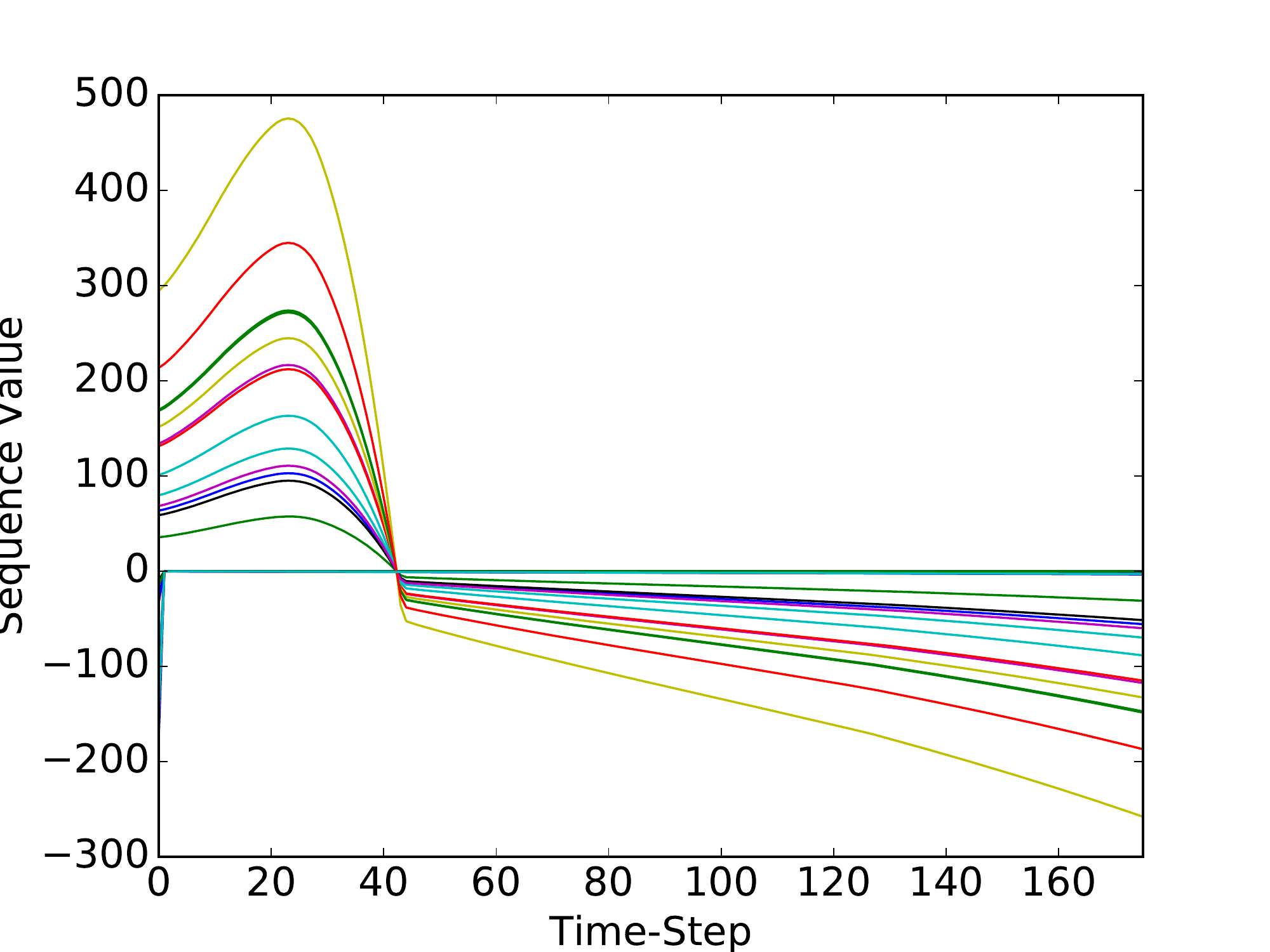}}
\caption{25 sequences generated from the ground truth SRU model.}
\label{fig:syn1d}
\end{center}
\vskip -0.3in
\end{figure}

After the ground truth SRU was constructed we generated the training, validation, and testing sequences. As can be seen in Figure \ref{fig:syn1d}, the sequences follow a simple pattern: at the start negative values are quickly pushed to zero and positive values follow a parabolic line until hitting zero, at which point they slope downward depending on initial values. While simple, it is clear that trained recurrent units must be able to hold long-term information since all sequences converge at one point and future behaviour depends on initial values.

We look to minimize the mean of squared errors (MSE); that is, the loss we consider per sequence is $ \frac{1}{175} \sum_{t=1}^{175} | x_{t+1} - p_{t} |^2 $, where $p_t$ is the output of the network after being fed $x_t$. We conducted 100 trials of Bayesian optimization as described above and obtained the following results in Table \ref{tbl:syn1d}.
\begin{table}[h]
\vskip -0.1in
\caption{MSEs for synthetically generated dataset.}
\label{tbl:syn1d}
\vskip 0.1in
\begin{center}
\begin{small}
\begin{tabular}{cccc}
\hline
\abovespace\belowspace
& SRU & GRU & LSTM \\  
\hline
\abovespace \belowspace
Error &\textbf{0.62} & 21.72 & 161.62 \\
\hline
\end{tabular}
\end{small}
\end{center}
\vskip -0.1in
\end{table}

Not surprisingly, the SRU performs far better than traditional gated recurrent units. This suggests that the types of long-term statistical relationships captured by the SRU are indeed different than those of traditional recurrent units. As previously mentioned, the SRU is able to obtain a multitude of different views from its statistics, a task that traditional units achieve less efficiently since they must devote one whole memory cell per viewpoint and statistic.
As we show below, the SRU is able to outperform traditional gated units in long term problems even for real data that is not generated from its model class.

\subsection{MNIST Image Classification}
\begin{figure}[ht]
\vskip -0.0in
\begin{center}
    \raisebox{0.5\height}{\includegraphics[width=0.25\columnwidth]{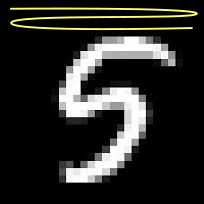}}\quad
    \includegraphics[width=0.66\columnwidth]{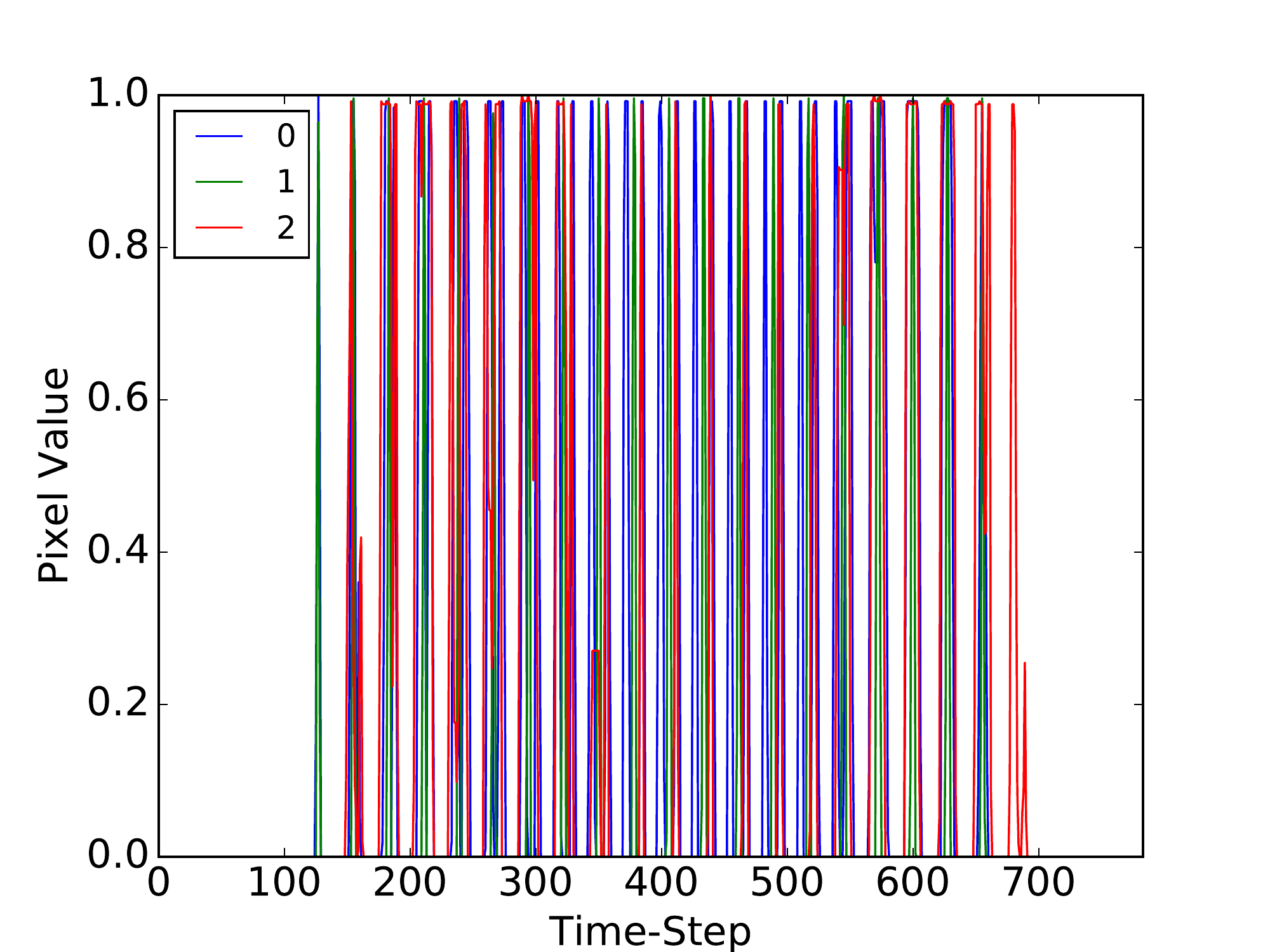}
\caption{Right: example MNIST $28\times 28$ image, which is taken as a pixel-by-pixel sequence of length $784$ unrolled as shown in yellow. Left: example pixel sequences for 0, 1, and 2 digit images.}
\label{fig:mnist_img}
\end{center}
\vskip -0.0in
\end{figure}

Next we explore the ability of recurrent units to use long-term dependencies in ones data with a synthetic task using a real dataset. It has been observed that LSTMs perform poorly in classifying a long pixel-by-pixel sequence of MNIST digits \cite{le2015simple}. In this synthetic task, each $28 \times 28$ gray-scale MNIST digit image is flattened and observed as a sequence $\{ x_1, \ldots, x_{784} \}$, where $x_i \in [0, 1]$ (see Figure \ref{fig:mnist_img}). The task is, based on the output observed after feeding $x_{784}$ through the network, to classify the digit of the corresponding image in $\{0, \ldots, 9\}$. Hence, we project the output after $x_{784}$ of each recurrent unit to $10$ dimensions and use a softmax activation. 

We report the Bayesian optimized results below in Table \ref{tbl:mnist}; due to resource constraints each trial consisted only of 10K training iterations. We see that the SRU is able to out-perform both GRUs and LSTMs. Given the long length and dependencies of pixel sequences in this experiment, it is not surprising that SRUs' abilities to capture long-term dependencies are aiding it to achieve a much lower error.

\begin{table}[th]
\vskip -0.1in
\caption{Test error rate for MNIST pixel sequence classification.}
\label{tbl:mnist}
\vskip 0.0in
\begin{center}
\begin{small}
\begin{tabular}{cccc}
\hline
\abovespace\belowspace
& SRU & GRU & LSTM \\  
\hline
\abovespace \belowspace
Error Rate &\textbf{0.11} & 0.28 & 0.48 \\
\hline
\end{tabular}
\end{small}
\end{center}
\vskip 0.0in
\end{table}

\subsubsection{Dissective Study}
Next, we study the behavior of the statistical recurrent unit with a dissective study where we vary several parameters of the architecture. 
We consider variants to the base model with:
\texttt{num\_stats=200}; \texttt{r\_dims=60}; \texttt{num\_units=200}.
We keep the parameters \texttt{initial\_learning\_rate}, \texttt{lr\_decay} fixed at the the optimal values found ($0.1$, $0.99$ respectively) unless we find no learning, in which case we also try learning rates of $0.01$ and $0.001$.

\paragraph{The need for multi-scaled recurrent statistics.}
Recall that we designed the statistics used by the SRU expressly 
to capture long term time dependencies in sequences. We did so both with recurrent statistics, i.e. statistics that themselves depend on previous points' statistics, and with multi-scaled averages. We show below that both of these time-dependent design choices are vital to capturing long term dependencies in data. Furthermore, we show that the use of ReLU statistics lends itself to better learning.

We explored the impact that time-dependent statistics had on learning by first considering naive i.i.d. summary statistics for sequences. This was achieved by using \texttt{r\_dims=0} and $\alpha \in A=\{0.99999\}$. Here no past-dependent context is used for statistics, i.e. we used i.i.d.-type statistics as is typical for unordered sets. Furthermore, the use of a single scale $\alpha$ near $1$ means that all of the points' statistics will be weighted nearly identically \eqref{eq:unrolled} regardless of index. We optimized the SRU when using no recurrent statistics and a single scale (\texttt{iid}), when using recurrent statistics with a single scale (\texttt{recur}), and when using no recurrent statistics with multiple scales (\texttt{multi}). We report errors below in Table \ref{tbl:mnist_iidstats}.

\begin{table}[th]
\vskip -0.1in
\caption{Test error rate for MNIST pixel sequence classification.}
\label{tbl:mnist_iidstats}
\vskip 0.0in
\begin{center}
\begin{small}
\begin{tabular}{cccc}
\hline
\abovespace\belowspace
& \texttt{iid} & \texttt{recur} & \texttt{multi} \\  
\hline
\abovespace \belowspace
Error Rate & 0.88 & 0.88 & 0.63 \\
\hline
\end{tabular}
\end{small}
\end{center}
\vskip 0.0in
\end{table}

Predictably, we cannot learn by simply keeping i.i.d. type statistics of pixel values at a single scale. Furthermore, we find that only using recurrent statistics (\texttt{recur}) in the SRU is not enough. It is interesting to note, however, that keeping i.i.d. statistics at multiple scales is able to predict digits with limited success. This lends evidence for the need of \emph{both} recurrent statistics and multiple scales.

Next, we explored the effects of the scales at which we keep our statistics by varying from $\alpha\in A = \{ 0.0, 0.5, 0.9, 0.99, 0.999 \}$ considering $\alpha\in A = \{ 0.0, 0.5, 0.9 \}$, $\alpha\in A = \{ 0.0, 0.5, 0.9, 0.99 \}$. We see in Table \ref{tbl:mnist_alphas} that additional, longer scales aid our learning for this dataset. This is not very surprising given the long term nature of the pixel sequences.

\begin{table}[th]
\vskip -0.1in
\caption{Test error rate for MNIST pixel sequence classification.}
\label{tbl:mnist_alphas}
\vskip 0.0in
\begin{center}
\begin{small}
\begin{tabular}{cccc}
\hline
\abovespace\belowspace
$A$ & \begin{tiny}$\{ 0.0, 0.5, 0.9 \}$\end{tiny} & \begin{tiny}$\{ 0.0, 0.5, 0.9, 0.99 \}$\end{tiny} &  \\  
\hline
\abovespace \belowspace
Error Rate & 0.79 & 0.21 &  \\
\hline
\end{tabular}
\end{small}
\end{center}
\vskip 0.0in
\end{table}

Lastly, we considered the use of non-ReLU statistics by changing the element-wise non-linearity $f(\cdot)$ \eqref{eq:recur_stats}-\eqref{eq:out} to be the hyperbolic tangent $f(\cdot) = \tanh(\cdot)$. We postulated that the use of ReLUs would help our learning since they have been observed to better handle the problem of vanishing gradients. We find evidence of this when swapping ReLUs for hyperbolic tangent units in SRUs: we get an error rate of $0.18$ when using hyperbolic tangent units. Although the previous uses of ReLUs in RNN required careful initialization \cite{le2015simple}, SRUs are able to use ReLUs for better learning without an special considerations.

\paragraph{Dimension of recurrent summary.}
Next we explore the affect of varying the number of dimensions used for the recurrent summary of statistics $r_t$ \eqref{eq:recur_stats}. We consider \texttt{r\_dims} in $\{5, 20, 240\}$. As previously discussed $r_t$ provides a context based on past data so that the SRU may produce non-i.i.d. statistics as it moves along a sequences. As one would expect the dimensionality of $r_t$ will limit the information flow from the past and values that are too small will hinder performance. It is also interesting to see that after enough dimensions, there are diminishing returns to adding more.

\begin{table}[th]
\vskip -0.1in
\caption{Test error rate varying recurrent summary $r_t$.}
\label{tbl:mnist_rdims}
\vskip 0.1in
\begin{center}
\begin{small}
\begin{tabular}{cccc}
\hline
\abovespace\belowspace
\texttt{r\_dims} & 5 & 20 & 240  \\  
\hline
\abovespace \belowspace
Error Rate & 0.25 & 0.20 & 0.10 \\
\hline
\end{tabular}
\end{small}
\end{center}
\vskip 0.0in
\end{table}

\paragraph{Number of statistics and outputs.}
Finally, we vary the number of statistics \texttt{num\_stats}, and outputs \texttt{units}. Interestingly the SRU seems robust to the number of outputs propagated in the network. However, performance is considerably affected by the number of statistics considered. 

\begin{table}[h]
\vskip -0.1in
\caption{Test error rate varying number of units.}
\label{tbl:mnist_numstatsunits}
\vskip 0.0in
\begin{center}
\begin{small}
\begin{tabular}{ccccccc}
\hline
\abovespace
 & \multicolumn{2}{c}{\texttt{num\_stats}}  & \multicolumn{2}{c}{\texttt{units}} \\\belowspace
 & 10 & 50 & 10 & 50  \\  
\hline
\abovespace \belowspace
Error Rate & 0.88 & 0.32 &  0.15 & 0.15\\
\hline
\end{tabular}
\end{small}
\end{center}
\vskip 0.0in
\end{table}


\subsection{Polyphonic Music Modeling}
Henceforth we consider real data and sequence learning tasks.
First, we used the polyphonic music datasets from
\citet{boulanger2012modeling}. Each time-step
is a binary vector representing the notes played at the respective time-step. Since we were required to
predict binary vectors we used the element-wise sigmoid $\sigma$. I.e., the binary vector of notes $x_{t+1}$ was modeled as
$\sigma\left(p_t\right)$, where $p_t$ is the output after feeding 
$x_t$ (and previous values $x_1, \ldots, x_{t-1}$)
through the recurrent network.

It is interesting to note in Table \ref{tbl:poly} that the SRU is able to outperform one of the traditional gated units in every dataset and it outperforms both in two datasets.

\begin{table}[th]
\vskip -0.1in
\caption{Test negative log-likelihood for polyphonic music data.}
\label{tbl:poly}
\vskip 0.1in
\begin{center}
\begin{small}
\begin{tabular}{lcccr}
\hline
\abovespace\belowspace
Data set & SRU & GRU & LSTM \\
\hline
\abovespace
JSB    & \textbf{8.260} &  8.548 &  8.393 \\
Muse & 6.336 &  6.429 &  \textbf{6.293} \\
Nottingham    & 3.362 & 3.386 &  \textbf{3.359}  \\
\belowspace
Piano   & \textbf{7.737} &  7.929 &  7.931 \\
\hline
\end{tabular}
\end{small}
\end{center}
\vskip -0.1in
\end{table}


\subsection{Electronica-Genre Music MFCC}
In the following experiment we modeled the Mel frequency cepstrum coefficients (MFCCs) in a dataset of nearly $18\,000$ scraped 30s sound clips of electronica-genre songs. MFCCs are perceptually based spectral features positioned logarithmically on the mel scale, which approximates the human auditory system's response \cite{muller2007information}. We looked to model the 13 real-valued coefficients using the recurrent units, by modeling $x_{t+1}$ as a projection of the output of a recurrent unit after being fed 
$x_1, \ldots, x_{t}$.

\begin{table}[h]
\vskip -0.1in
\caption{Test-set MSEs of MFCC Music data.}
\label{tbl:mfcc}
\vskip 0.1in
\begin{center}
\begin{small}
\begin{tabular}{lcccr}
\hline
\abovespace\belowspace
 & SRU & GRU & LSTM \\
\hline
\abovespace
\belowspace
  Error & \textbf{1.176} &  2.080 &  1.183 \\
\hline
\end{tabular}
\end{small}
\end{center}
\vskip -0.1in
\end{table}

As can be seen in Table \ref{tbl:mfcc}, SRUs again are outperforming gated architectures and are especially beating GRUs by a wider margin.

\subsection{Climate Data}
Next we consider weather data prediction using the North America Regional Reanalysis (NARR) Project.
The dataset provides a long-term set of consistent climate data on a regional scale for the North American domain. The period of the reanalyses is from October 1978 to the present and analyses were made 8 times daily (3 hour intervals). 

\begin{figure}[h]
\vskip -0.0in
\begin{center}
    \includegraphics[width=0.49\columnwidth]{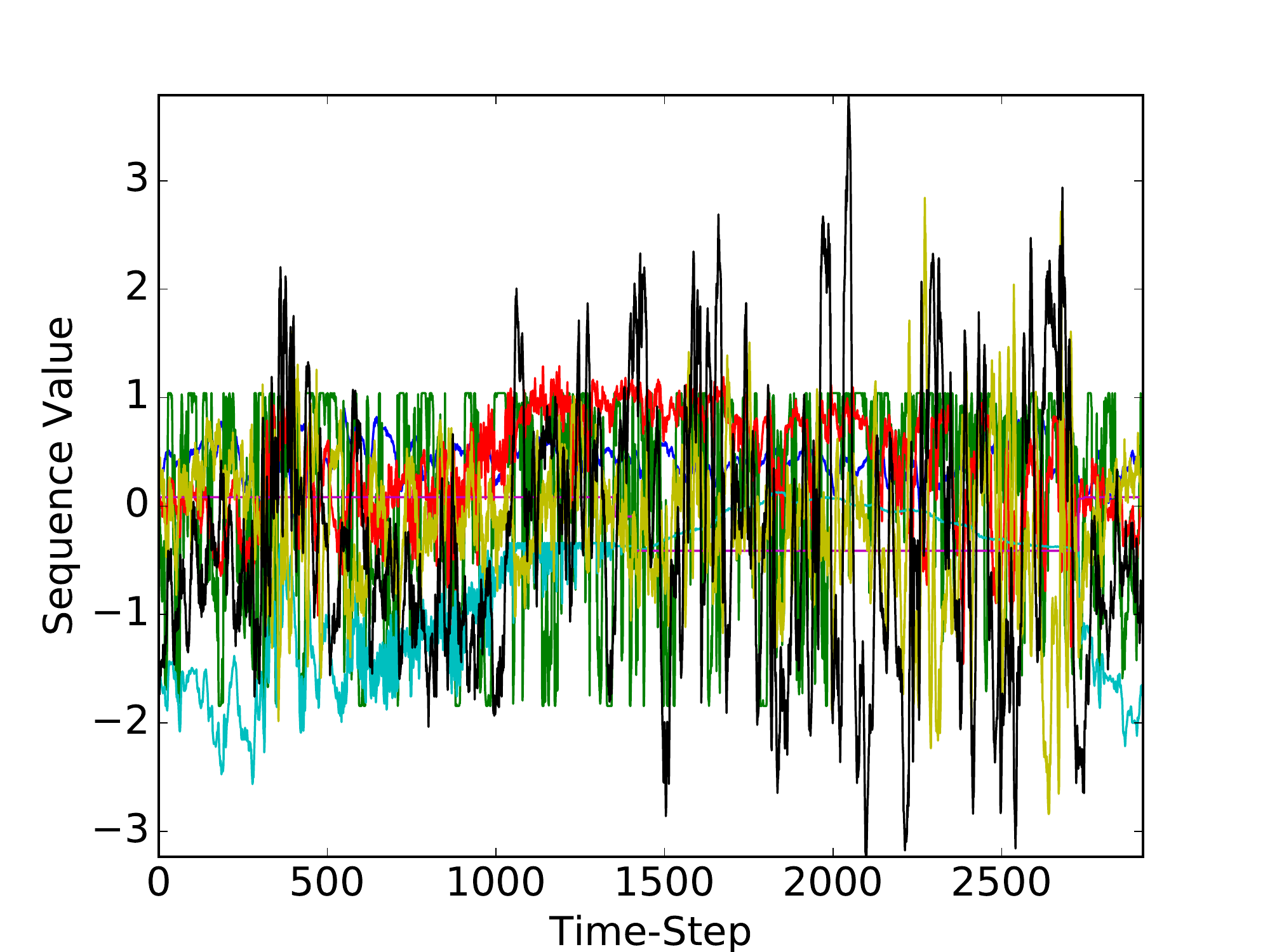}\ 
    \includegraphics[width=0.49\columnwidth]{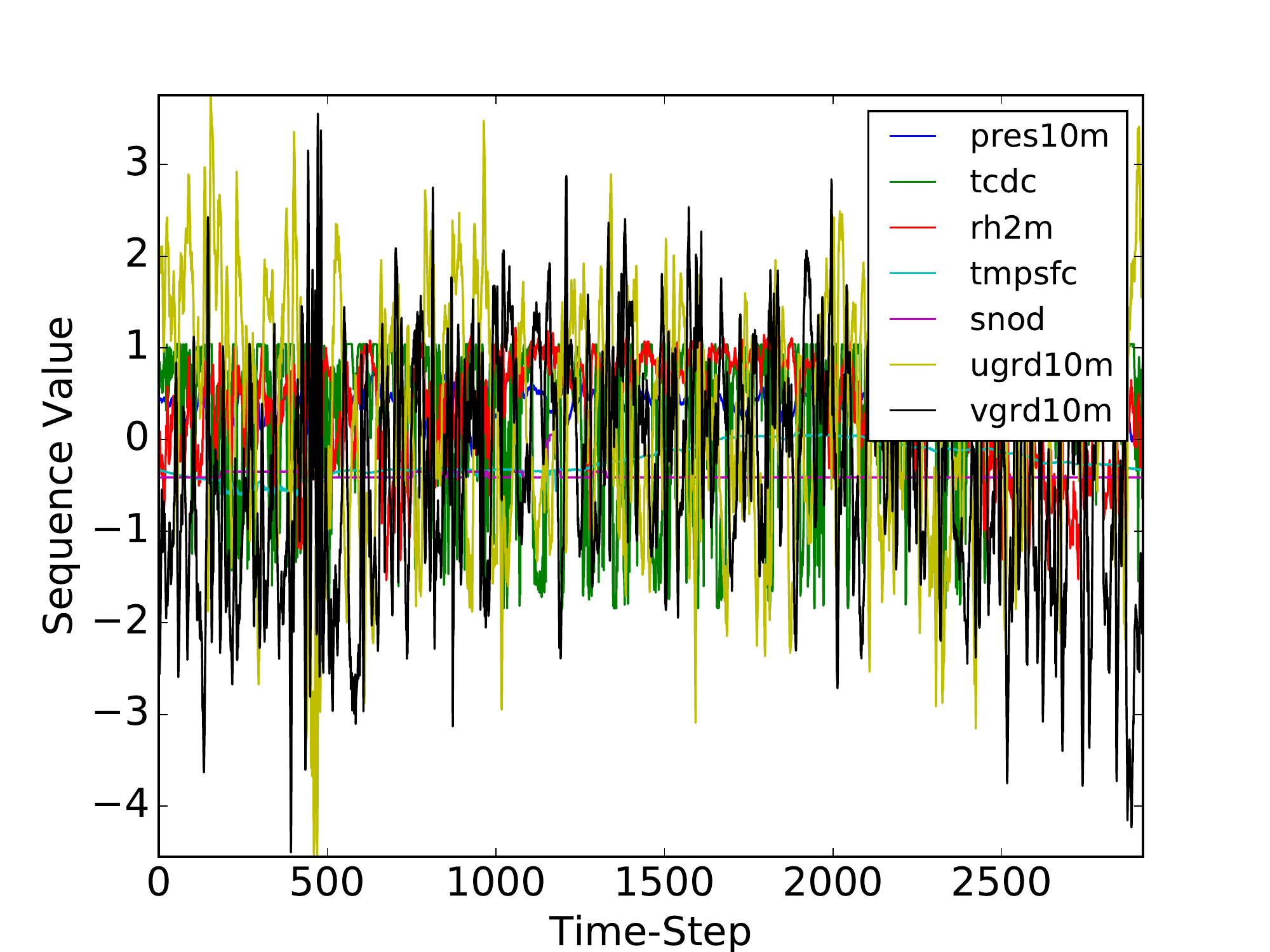}
\caption{Two example sequences for weather variables at distinct locations for the year 2006.}
\label{fig:weather}
\end{center}
\vskip -0.1in
\end{figure}

We take our input sequences to be year-long sequences of weather variables in a location for the year 2006. I.e. an input sequence will be a 2920 length sequence of weather variables at a given lat/lon coordinate.
We considered the following 7 variables: \texttt{pres10m}, 10 m pressure (pa); \texttt{tcdc}, total cloud cover (\%); \texttt{rh2m}, relative humidity 2m (\%); \texttt{tmpsfc}, surface temperature (k); \texttt{snod}, snow depth surface (m); \texttt{ugrd10m}, u component of wind 10m above ground; \texttt{vgrd10m}, v component of wind 10m above ground. The variables were standardized, see Figure \ref{fig:weather} for example sequences.

Below we see results using $51\,200$ training location sequences and $6\,400$ validation and testing instances. Again, we look to model the next point in a sequence as a projection of the output of the recurrent unit after feeding the previous points. One may see in Table \ref{tbl:weather} that SRUs and LSTMs perform nearly identically; perhaps the cyclical nature of climate data was beneficial to the gated units.

\begin{table}[th]
\vskip -0.1in
\caption{Test MSEs for weather data.}
\label{tbl:weather}
\vskip 0.1in
\begin{center}
\begin{small}
\begin{tabular}{lcccr}
\hline
\abovespace\belowspace
 & SRU & GRU & LSTM \\
\hline
\abovespace
\belowspace
 Error  & \textbf{0.465} &  0.487 &  0.466 \\
\hline
\end{tabular}
\end{small}
\end{center}
\vskip -0.1in
\end{table}

\subsection{SportVu NBA Tracking data}
Finally, we look to predict the positions of National Basketball Association (NBA) players based on previous court positions during a play. Optical tracking data for this project were provided by STATS LLC from their SportVU product and obtained from \cite{NBA}. The data are composed of $x$ and $y$ coordinates for each of the ten players and the ball. We again minimize the squared norm of errors for predictions.

\begin{figure}[ht]
\vskip -0.0in
\begin{center}
    \includegraphics[width=0.49\columnwidth]{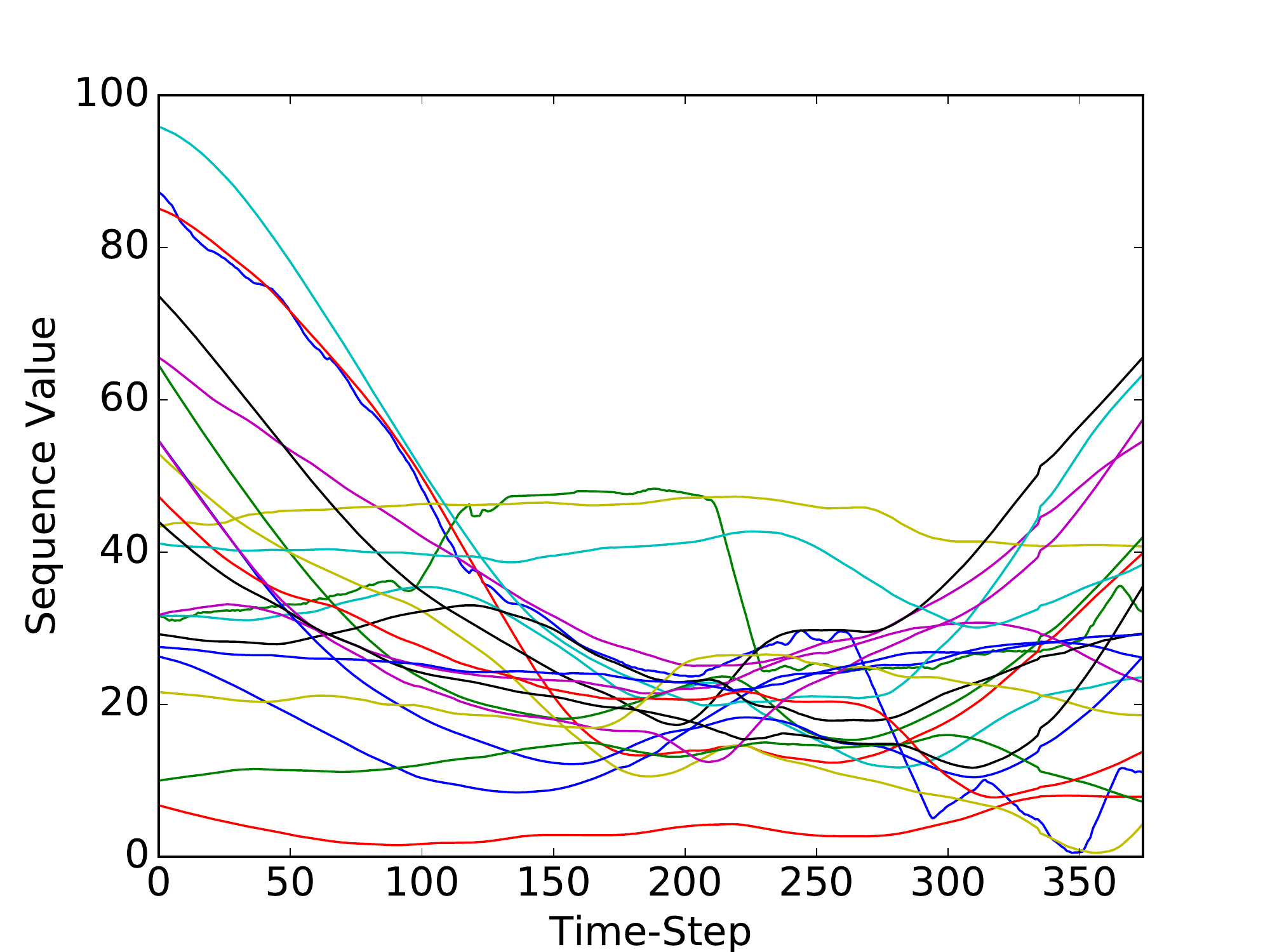}\ 
    \includegraphics[width=0.49\columnwidth]{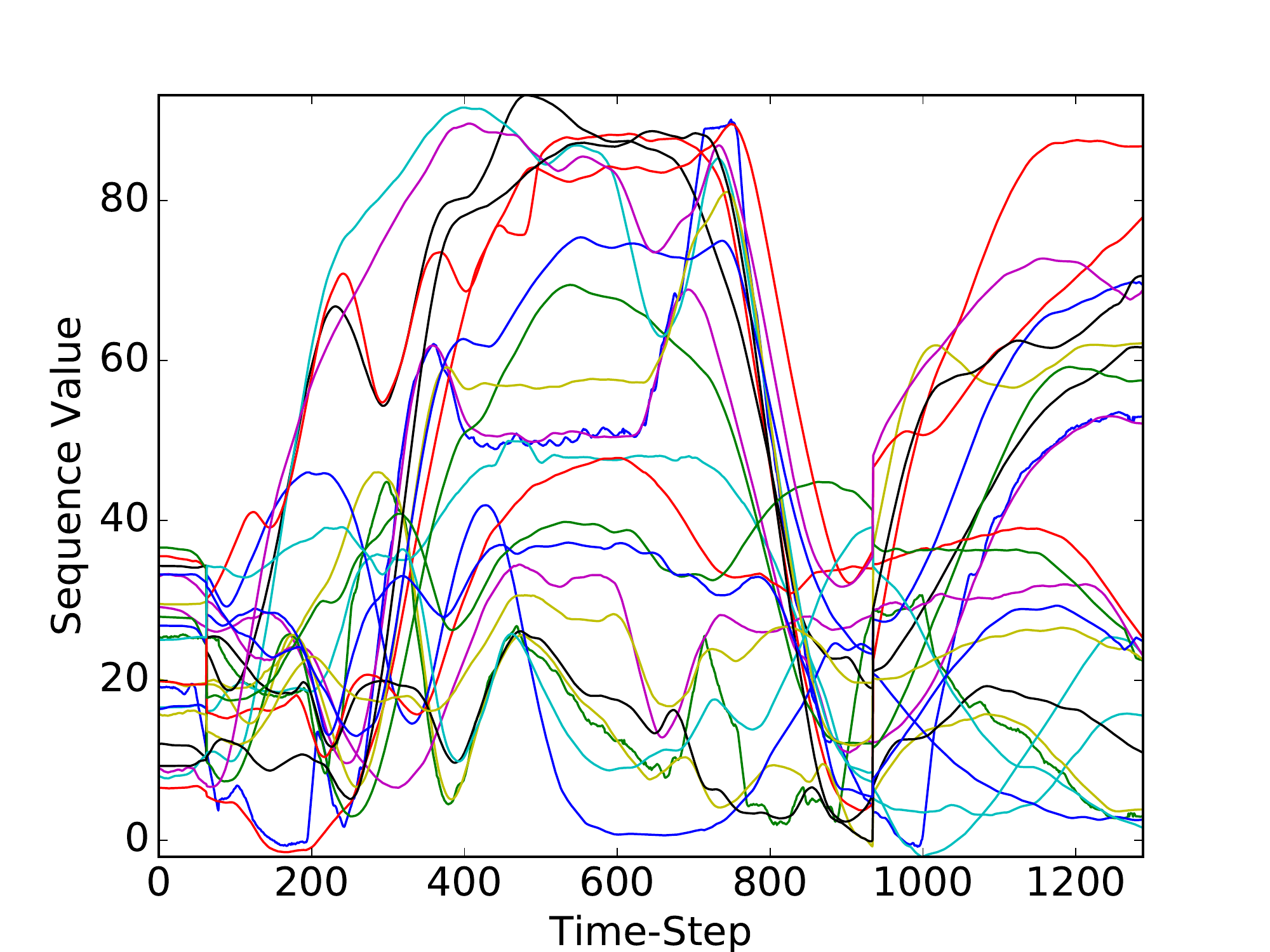}
\caption{Example player/ball $x, y$ positions for two plays.}
\label{fig:nba}
\end{center}
\vskip -0.1in
\end{figure} 

\begin{table}[h]
\vskip -0.1in
\caption{Test-set MSEs of NBA data.}
\label{tbl:nba}
\vskip 0.1in
\begin{center}
\begin{small}
\begin{tabular}{lcccr}
\hline
\abovespace\belowspace
 & SRU & GRU & LSTM \\
\hline
\abovespace
\belowspace
 Error  & \textbf{34.505} &  329.921 &  296.908 \\
\hline
\end{tabular}
\end{small}
\end{center}
\vskip -0.15in
\end{table}

We observed a large margin of improvement for SRUs over gated architectures in Table \ref{tbl:nba} that is reminiscent of the synthetic data experiment in $\S$\ref{sec:synth}. This suggests that this dataset contains long term dependencies that the SRU is able to exploit.

\section{Discussion}

We believe that the use of summary statistics has been under-explored in modern recurrent units. Although recent studies in convolutional networks have considered global average pooling, which is essentially using high-level summary statistics to represent images, there has been little exploration of summary statistics for modern recurrent networks. To this end we introduce the Statistical Recurrent Unit, a novel architecture that seeks to capture long term dependencies in data using only simple moving averages and rectified-linear units.

The SRU was motivated by the success of mean-map embeddings for representing unordered datasets, and may be interpreted as an alteration of MMEs for sequential data. The main modifications are as follows: first, the SRU uses data-driven statistics unlike typical MMEs, which will use RKHS features from an a-priori selected class of kernels; second, SRUs will use recurrent statistics that dependent not only on a current point, but on previous points' statistics through a condensation of kept moving averages; third, the SRU will keep moving averages at various scales. 
We provide evidence that the combination of these modifications yield much better results than any one of them in isolation.
The resulting recurrent unit is especially adept for capturing long term dependencies in data and readily has access to a combinatorial number of viewpoints of past windows through simple linear combinations.

We showed empirically that the SRU is better equipped that traditional gated units for long term dependencies via synthetic and real-world data experiments.

\pagebreak

\bibliography{example_paper}

\begin{thebibliography}{27}
\providecommand{\natexlab}[1]{#1}
\providecommand{\url}[1]{\texttt{#1}}
\expandafter\ifx\csname urlstyle\endcsname\relax
  \providecommand{\doi}[1]{doi: #1}\else
  \providecommand{\doi}{doi: \begingroup \urlstyle{rm}\Url}\fi

\bibitem[NBA()]{NBA}
Nba movement data.
\newblock \url{https://github.com/sealneaward/nba-movement-data}.
\newblock Accessed: 2016-10-17.

\bibitem[Abadi et~al.(2016)Abadi, Barham, Chen, Chen, Davis, Dean, Devin,
  Ghemawat, Irving, Isard, et~al.]{abadi2016tensorflow}
Abadi, Mart{\'\i}n, Barham, Paul, Chen, Jianmin, Chen, Zhifeng, Davis, Andy,
  Dean, Jeffrey, Devin, Matthieu, Ghemawat, Sanjay, Irving, Geoffrey, Isard,
  Michael, et~al.
\newblock Tensorflow: A system for large-scale machine learning.
\newblock In \emph{Proceedings of the 12th USENIX Symposium on Operating
  Systems Design and Implementation (OSDI). Savannah, Georgia, USA}, 2016.

\bibitem[Bergstra \& Bengio(2012)Bergstra and Bengio]{bergstra2012random}
Bergstra, James and Bengio, Yoshua.
\newblock Random search for hyper-parameter optimization.
\newblock \emph{Journal of Machine Learning Research}, 13\penalty0
  (Feb):\penalty0 281--305, 2012.

\bibitem[Bergstra et~al.(2015)Bergstra, Komer, Eliasmith, Yamins, and
  Cox]{bergstra2015hyperopt}
Bergstra, James, Komer, Brent, Eliasmith, Chris, Yamins, Dan, and Cox, David~D.
\newblock Hyperopt: a python library for model selection and hyperparameter
  optimization.
\newblock \emph{Computational Science \& Discovery}, 8\penalty0 (1):\penalty0
  014008, 2015.

\bibitem[Boulanger-Lewandowski et~al.(2012)Boulanger-Lewandowski, Bengio, and
  Vincent]{boulanger2012modeling}
Boulanger-Lewandowski, Nicolas, Bengio, Yoshua, and Vincent, Pascal.
\newblock Modeling temporal dependencies in high-dimensional sequences:
  Application to polyphonic music generation and transcription.
\newblock \emph{arXiv preprint arXiv:1206.6392}, 2012.

\bibitem[Cho et~al.(2014)Cho, Van~Merri{\"e}nboer, Bahdanau, and
  Bengio]{cho2014properties}
Cho, Kyunghyun, Van~Merri{\"e}nboer, Bart, Bahdanau, Dzmitry, and Bengio,
  Yoshua.
\newblock On the properties of neural machine translation: Encoder-decoder
  approaches.
\newblock \emph{arXiv preprint arXiv:1409.1259}, 2014.

\bibitem[Chung et~al.(2014)Chung, Gulcehre, Cho, and
  Bengio]{chung2014empirical}
Chung, Junyoung, Gulcehre, Caglar, Cho, KyungHyun, and Bengio, Yoshua.
\newblock Empirical evaluation of gated recurrent neural networks on sequence
  modeling.
\newblock \emph{arXiv preprint arXiv:1412.3555}, 2014.

\bibitem[Donahue et~al.(2015)Donahue, Anne~Hendricks, Guadarrama, Rohrbach,
  Venugopalan, Saenko, and Darrell]{donahue2015long}
Donahue, Jeffrey, Anne~Hendricks, Lisa, Guadarrama, Sergio, Rohrbach, Marcus,
  Venugopalan, Subhashini, Saenko, Kate, and Darrell, Trevor.
\newblock Long-term recurrent convolutional networks for visual recognition and
  description.
\newblock In \emph{Proceedings of the IEEE conference on computer vision and
  pattern recognition}, pp.\  2625--2634, 2015.

\bibitem[Edwards \& Storkey(2016)Edwards and Storkey]{edwards2016towards}
Edwards, Harrison and Storkey, Amos.
\newblock Towards a neural statistician.
\newblock \emph{arXiv preprint arXiv:1606.02185}, 2016.

\bibitem[Elman(1990)]{elman1990finding}
Elman, Jeffrey~L.
\newblock Finding structure in time.
\newblock \emph{Cognitive science}, 14\penalty0 (2):\penalty0 179--211, 1990.

\bibitem[Graves \& Jaitly(2014)Graves and Jaitly]{graves2014towards}
Graves, Alex and Jaitly, Navdeep.
\newblock Towards end-to-end speech recognition with recurrent neural networks.
\newblock In \emph{ICML}, volume~14, pp.\  1764--1772, 2014.

\bibitem[Graves et~al.(2013)Graves, Mohamed, and Hinton]{graves2013speech}
Graves, Alex, Mohamed, Abdel-rahman, and Hinton, Geoffrey.
\newblock Speech recognition with deep recurrent neural networks.
\newblock In \emph{Acoustics, speech and signal processing (icassp), 2013 ieee
  international conference on}, pp.\  6645--6649. IEEE, 2013.

\bibitem[Hochreiter \& Schmidhuber(1997)Hochreiter and
  Schmidhuber]{hochreiter1997long}
Hochreiter, Sepp and Schmidhuber, J{\"u}rgen.
\newblock Long short-term memory.
\newblock \emph{Neural computation}, 9\penalty0 (8):\penalty0 1735--1780, 1997.

\bibitem[Hochreiter et~al.(2001)Hochreiter, Bengio, Frasconi, and
  Schmidhuber]{hochreiter2001gradient}
Hochreiter, Sepp, Bengio, Yoshua, Frasconi, Paolo, and Schmidhuber, J{\"u}rgen.
\newblock Gradient flow in recurrent nets: the difficulty of learning long-term
  dependencies, 2001.

\bibitem[Iandola et~al.(2016)Iandola, Han, Moskewicz, Ashraf, Dally, and
  Keutzer]{iandola2016squeezenet}
Iandola, Forrest~N, Han, Song, Moskewicz, Matthew~W, Ashraf, Khalid, Dally,
  William~J, and Keutzer, Kurt.
\newblock Squeezenet: Alexnet-level accuracy with 50x fewer parameters and< 0.5
  mb model size.
\newblock \emph{arXiv preprint arXiv:1602.07360}, 2016.

\bibitem[Jarrett et~al.(2009)Jarrett, Kavukcuoglu, LeCun,
  et~al.]{jarrett2009best}
Jarrett, Kevin, Kavukcuoglu, Koray, LeCun, Yann, et~al.
\newblock What is the best multi-stage architecture for object recognition?
\newblock In \emph{Computer Vision, 2009 IEEE 12th International Conference
  on}, pp.\  2146--2153. IEEE, 2009.

\bibitem[Le et~al.(2015)Le, Jaitly, and Hinton]{le2015simple}
Le, Quoc~V, Jaitly, Navdeep, and Hinton, Geoffrey~E.
\newblock A simple way to initialize recurrent networks of rectified linear
  units.
\newblock \emph{arXiv preprint arXiv:1504.00941}, 2015.

\bibitem[Lin et~al.(2013)Lin, Chen, and Yan]{lin2013network}
Lin, Min, Chen, Qiang, and Yan, Shuicheng.
\newblock Network in network.
\newblock \emph{arXiv preprint arXiv:1312.4400}, 2013.

\bibitem[Mikolov et~al.(2015)Mikolov, Joulin, Chopra, Mathieu, and
  Ranzato]{mikolov2014learning}
Mikolov, Tomas, Joulin, Armand, Chopra, Sumit, Mathieu, Michael, and Ranzato,
  Marc'Aurelio.
\newblock Learning longer memory in recurrent neural networks.
\newblock \emph{arXiv preprint arXiv:1412.7753}, 2015.

\bibitem[Muandet et~al.(2016)Muandet, Fukumizu, Sriperumbudur, and
  Sch{\"o}lkopf]{muandet2016kernel}
Muandet, Krikamol, Fukumizu, Kenji, Sriperumbudur, Bharath, and Sch{\"o}lkopf,
  Bernhard.
\newblock Kernel mean embedding of distributions: A review and beyonds.
\newblock \emph{arXiv preprint arXiv:1605.09522}, 2016.

\bibitem[M{\"u}ller(2007)]{muller2007information}
M{\"u}ller, Meinard.
\newblock \emph{Information retrieval for music and motion}, volume~2.
\newblock Springer, 2007.

\bibitem[Nair \& Hinton(2010)Nair and Hinton]{nair2010rectified}
Nair, Vinod and Hinton, Geoffrey~E.
\newblock Rectified linear units improve restricted boltzmann machines.
\newblock In \emph{Proceedings of the 27th international conference on machine
  learning (ICML-10)}, pp.\  807--814, 2010.

\bibitem[Pascanu et~al.(2013)Pascanu, Mikolov, and
  Bengio]{pascanu2013difficulty}
Pascanu, Razvan, Mikolov, Tomas, and Bengio, Yoshua.
\newblock On the difficulty of training recurrent neural networks.
\newblock \emph{ICML (3)}, 28:\penalty0 1310--1318, 2013.

\bibitem[Ramsay \& Silverman(2002)Ramsay and Silverman]{ramsay2002applied}
Ramsay, J.O. and Silverman, B.W.
\newblock \emph{Applied functional data analysis: methods and case studies},
  volume~77.
\newblock Springer New York:, 2002.

\bibitem[Smola et~al.(2007)Smola, Gretton, Song, and
  Sch{\"o}lkopf]{smola2007hilbert}
Smola, Alex, Gretton, Arthur, Song, Le, and Sch{\"o}lkopf, Bernhard.
\newblock A hilbert space embedding for distributions.
\newblock In \emph{International Conference on Algorithmic Learning Theory},
  pp.\  13--31. Springer, 2007.

\bibitem[Vinyals et~al.(2015)Vinyals, Kaiser, Koo, Petrov, Sutskever, and
  Hinton]{vinyals2015grammar}
Vinyals, Oriol, Kaiser, {\L}ukasz, Koo, Terry, Petrov, Slav, Sutskever, Ilya,
  and Hinton, Geoffrey.
\newblock Grammar as a foreign language.
\newblock In \emph{Advances in Neural Information Processing Systems}, pp.\
  2773--2781, 2015.

\bibitem[Zaremba et~al.(2014)Zaremba, Sutskever, and
  Vinyals]{zaremba2014recurrent}
Zaremba, Wojciech, Sutskever, Ilya, and Vinyals, Oriol.
\newblock Recurrent neural network regularization.
\newblock \emph{arXiv preprint arXiv:1409.2329}, 2014.

\end{thebibliography}
\bibliographystyle{icml2017}

\end{document}